\documentclass[runningheads]{llncs}

\usepackage{graphicx}
\usepackage{amsfonts}
\usepackage{amssymb}
\usepackage{xcolor}
\usepackage{amsmath}
\usepackage{csquotes}
\usepackage[linesnumbered]{algorithm2e}
\SetAlCapSty{}
\SetAlgoInsideSkip{bigskip}
\usepackage{hyperref}

\newcommand{\labs}{\left|}
\newcommand{\rabs}{\right|}

\begin{document}

\title{Adjusted Measures for Feature Selection Stability for Data Sets with Similar Features}
\titlerunning{Adjusted Stability Measures}

\author{Andrea Bommert \and J\"org Rahnenf\"uhrer}

\institute{TU Dortmund University, 44221 Dortmund, Germany \\ \email{bommert@statistik.tu-dortmund.de}\footnote{The source code for the experiments and analyses of this article is publicly available at \url{https://github.com/bommert/adjusted-stability-measures}.}}

\maketitle

\begin{abstract}
For data sets with similar features, for example highly correlated features, most existing stability measures behave in an undesired way: They consider features that are almost identical but have different identifiers as different features.
Existing adjusted stability measures, that is, stability measures that take into account the similarities between features, have major theoretical drawbacks.
We introduce new adjusted stability measures that overcome these drawbacks.
We compare them to each other and to existing stability measures based on both artificial and real sets of selected features.
Based on the results, we suggest using one new stability measure that considers highly similar features as exchangeable.
\end{abstract}

\section{Introduction}
Feature selection is one of the most fundamental problems in data analysis, machine learning, and data mining.
Recently, it has drawn increasing attention due to high-dimensional data sets emerging from many different fields.
Especially in domains where the chosen features are subject to further experimental research, the stability of the feature selection is very important.
Stable feature selection means that the set of selected features is robust with respect to different data sets from the same data generating distribution~\cite{kalousis2007stability}.
If for data sets from the same data generating process, very different sets of features are chosen, this questions not only the reliability of resulting models but could also lead to unnecessary expensive experimental research.

The evaluation of feature selection stability is an active area of research.
Overviews of existing stability measures are given in~\cite{he2010stable} and~\cite{lausser2013measuring}.
The theoretical properties of different stability measures are studied in~\cite{nogueira2018quantifying}.
An extensive empirical comparison of stability measures is given in~\cite{bommert2017multicriteria}.
The research that has been done in various aspects related to stability assessment is reviewed in~\cite{awada2012review}.

For data sets with similar features, the evaluation of feature selection stability is more difficult.
An example for such data sets are gene expression data sets, where genes of the same biological processes are often highly positively correlated.
The commonly used stability measures consider features that are almost identical but have different identifiers as different features.
Only little research has been performed concerning the assessment of feature selection stability for data sets with similar features.
Stability measures that take into account the similarities between features are defined in~\cite{yu2012stable}, \cite{zhang2009evaluating} and~\cite{zucknick2008comparing}.
These measures, however, have major theoretical drawbacks.
We call stability measures that consider the similarities between features \enquote{adjusted} stability measures.

In this paper,  we introduce new adjusted stability measures.
On both artificial and real sets of selected features, we compare them to each other and to existing stability measures and analyze their properties.
The remainder of this paper is organized as follows:
In Section~\ref{sec.methods}, the concept of feature selection stability is explained in detail and adjusted stability measures are defined.
The stability measures are compared in Section~\ref{sec.experiments.results}.
Section~\ref{sec.conclusions}~contains a summary of the findings and concluding remarks.

\section{Concepts and Methods}\label{sec.methods}
In Subsection~\ref{sec.stability}, feature selection stability is explained and in Subsection~\ref{sec.measures}, measures for quantifying feature selection stability are introduced.

\subsection{Feature Selection Stability}\label{sec.stability}
The stability of a feature selection algorithm is defined as the robustness of the set of selected features to different data sets from the same data generating distribution~\cite{kalousis2007stability}.
Stability quantifies how different training data sets affect the sets of chosen features.
For similar data sets, a stable feature selection algorithm selects similar sets of features.
An example for similar data sets could be data coming from different studies measuring the same features, possibly conducted at different places and times, as long as the assumption of the same underlying distribution is valid.

A lack of stability has three main reasons: too few observations, highly similar features and equivalent sets of features.
Consider a group of data sets, for which the number of observations does not greatly exceed the number of features, from the same data generating process. 
The subsets of features with maximal predictive quality on the respective data sets often differ between these data sets.
One reason is that there are features that seem beneficial for prediction, but that only help on the specific data set and not on new data from the same process.
Selecting such features and including them in a predictive model typically causes over-fitting.
Another reason is that there are features with similar predictive quality even though they are unrelated with respect to their content.
Due to the small number of observations, chance has a large influence on which of these features has the highest predictive quality on each data set.
The instability of feature selection resulting from both reasons is undesirable.

Regarding the case of highly similar and therefore almost identical features, it is likely that for some data sets, one feature is selected and for other data sets from the same process, another one of the similar features is chosen.
As the features are almost identical, it makes sense to label this as stable because the feature selection algorithm always chooses a feature with the same information.
Therefore, it is desirable to have a stability measure that takes into account the reason for the differences in the sets of chosen features.
However, most existing stability measures treat both situations equally: if the identifiers of the chosen features are different, the feature selection is rated unstable.

Regarding the case of equivalent feature sets, for some data sets, there are different sets of features that contain exactly the same information.
Finding all equivalent optimal subsets of features is an active field of research, see for example \cite{statnikov2013algorithms}, and worst-case intractable.
The selection of equivalent subsets of features is evaluated as unstable by all existing stability measures. 
Creating stability measures that can recognize equivalent sets of features is out of the scope of this paper.

\subsection{Adjusted Stability Measures}\label{sec.measures}
For the definition of the stability measures, the following notation is used:
Assume that there is a data generating process that generates observations of the $p$~features $X_1, \ldots, X_p$.
Further, assume that there are $m$~data sets that are generated by this process.
A feature selection method is applied to all data sets.
Let $V_i \subseteq \{X_1, \ldots, X_p\}$, \mbox{$i = 1, \ldots, m,$} denote the set of chosen features for the $i$-th data set and $\labs V_i \rabs$ the cardinality of this set.
The feature selection stability is assessed based on the similarity of the sets $V_1, \ldots, V_m$.
For all stability measures, large values correspond to high stability and small values  to low stability.

Many existing stability measures that do not consider similarities between features assess the stability based on the pairwise scores $\labs V_i \cap V_j \rabs$, see for example~\cite{bommert2017multicriteria} and~\cite{nogueira2018quantifying}.
An example for an unadjusted stability measure is
\[\text{SMU} = \frac{2}{m(m-1)}\sum\limits_{i=1}^{m-1} \sum\limits_{j=i+1}^{m} \frac{ \labs V_i \cap V_j \rabs - \frac{\labs V_i \rabs \cdot \labs V_j \rabs}{p}}{\sqrt{\labs V_i \rabs \cdot \labs V_j \rabs} - \frac{\labs V_i \rabs \cdot \labs V_j \rabs}{p}}.\]
$\frac{\labs V_i \rabs \cdot \labs V_j \rabs}{p}$ is the expected value of $\labs V_i \cap V_j \rabs$ if $\labs V_i \rabs$ and $\labs  V_j \rabs$ features are chosen at random with equal selection probabilities. 
$\sqrt{\labs V_i \rabs \cdot \labs V_j \rabs}$ is an upper bound for $\labs V_i \cap V_j \rabs$. 
Including it in the denominator makes 1 the maximum value of SMU.
If many of the sets $V_i$ and $V_j$ have a large overlap, the feature selection is evaluated as rather stable.
The basic idea of adjusted stability measures is to adjust the scores $\labs V_i \cap V_j \rabs$ in a way that different but highly similar features count towards stability.
Note that all of the following adjusted stability measures depend on a threshold $\theta$.
This threshold indicates how similar features have to be in order to be seen as exchangeable for stability assessment.

Zucknick et al.~\cite{zucknick2008comparing} extend the well known Jaccard index~\cite{jaccard1901etude}, considering the correlations between the features:
\begin{align*}
&\text{SMZ} = \frac{2}{m(m-1)}\sum\limits_{i=1}^{m-1} \sum\limits_{j=i+1}^{m} \frac{\labs V_i \cap V_j\rabs + C(V_i, V_j) + C(V_j, V_i)}{\labs V_i \cup V_j \rabs}  \quad  \text{with}\\
&C(V_i, V_j) = \sum\limits_{x \in V_i} \frac{1}{\labs V_j \rabs} \sum\limits_{y \in V_j \setminus V_i} \labs \text{Cor}(x,y) \rabs \, \mathbb{I}_{[\theta, \infty)} \left(\labs \text{Cor}(x,y) \rabs \right).
\end{align*}
$\labs \text{Cor}(x,y) \rabs$ is the absolute Pearson correlation between $x$ and $y$, $\theta \in [0,1]$ is a threshold, and $\mathbb{I}_S$ denotes the indicator function for a set $S$.
One could generalize this stability measure by allowing arbitrary similarity values from the interval $[0,1]$ instead of the absolute correlations.
A major drawback of this stability measure is that it is not corrected for chance.
Correction for chance~\cite{nogueira2018quantifying} means that the expected value of the stability measure for a random feature selection with equal selection probabilities for all features does not depend on the number of chosen features.

Zhang et al.~\cite{zhang2009evaluating} also present adjusted stability measures.
Their scores are developed for the comparison of two gene lists.
The scores they define are
\[ \text{nPOGR}_{ij} = \frac{K + O_{ij} - E\left[K + O_{ij}\right]}{\labs V_i \rabs - E\left[K + O_{ij}\right]} \]
with $ij \in \{12, 21\}$.
$K$ is defined as the number of genes that are included in both lists and regulated in the same direction.
$O_{ij}$ denotes the number of genes in list $i$ that are not in list $j$ but significantly positively correlated with at least one gene in list $j$.
For each pair of gene lists, two stability scores are obtained.

Yu et al. \cite{yu2012stable} combine the two scores $\text{nPOGR}_{ij}$ and $\text{nPOGR}_{ji}$ into one score for the special case $\labs V_i \rabs = \labs V_j \rabs$:
\[\text{nPOGR} = \frac{K + \frac{O_{ij} + O_{ji}}{2} - E\left[K + \frac{O_{ij} + O_{ji}}{2}\right]}{\labs V_i \rabs - E\left[K + \frac{O_{ij} + O_{ji}}{2}\right]}.\]

In this paper, we generalize this score to be applicable in the general context of feature selection with arbitrary feature sets $V_1, \ldots, V_m$ by
\begin{enumerate}
\item replacing the quantity $K$ by $\labs V_i \cap V_j \rabs$.
\item allowing the similarities between the features to be assessed by an arbitrary similarity measure instead of only considering significantly positive correlations, that is, replacing $ \frac{O_{ij} + O_{ji}}{2}$ by  $\frac{A(V_i, V_j) + A(V_j, V_i)}{2}$ with $A$ defined below.
\item replacing $\labs V_i \rabs$, which is the maximum value of $K + \frac{O_{ij} + O_{ji}}{2}$, by $\frac{\labs V_i \rabs + \labs V_j \rabs}{2}$, the maximum value of $\labs V_i \cap V_j \rabs + \frac{A(V_i, V_j) + A(V_j, V_i)}{2}$.
\item calculating the average of the scores for all pairs $V_i$, $V_j$, $i < j$.
\end{enumerate}

As a result, the stability measure
\begin{align*}
&\text{SMY} = \frac{2}{m(m-1)}\sum\limits_{i=1}^{m-1} \sum\limits_{j=i+1}^{m} \frac{\text{S}_{\text{SMY}}(V_i, V_j) - E\left[ \text{S}_{\text{SMY}}(V_i, V_j) \right]}{\frac{\labs V_i \rabs + \labs V_j \rabs}{2} - E\left[ \text{S}_{\text{SMY}}(V_i, V_j) \right]}\\
&\text{with} \quad \text{S}_{\text{SMY}}(V_i, V_j) = \labs V_i \cap V_j \rabs + \frac{A(V_i, V_j) + A(V_j, V_i)}{2}\\
&\text{and} \quad A(V_i, V_j) = \labs \{ x \in (V_i \setminus V_j) : \exists y \in (V_j \setminus V_i) \text{ with similarity}(x, y) \geq \theta\} \rabs
\end{align*}
is obtained.
$E$ denotes the expected value for a random feature selection and can be assessed in the same way as described below for SMA.
Similarity$(x,y) \in [0, 1]$ quantifies the similarity of the two features $x$ and $y$ and $\theta \in [0, 1]$ is a threshold.

In situations where $V_i$ and $V_j$ greatly differ in size and contain many similar features, the value of SMY may be misleading.
Consider a scenario with $\labs V_i \rabs \gg \labs V_j \rabs$, $\labs V_i \cap V_j \rabs = 0$, $A(V_i, V_j) = \labs V_i \rabs$, and $A(V_j, V_i) = \labs V_j \rabs$.
In such situations, there are many features in the larger set that are similar to the same feature in the smaller set.
Even though the sets $V_i$ and $V_j$ greatly differ with respect to feature redundancy and resulting effects for model building such as over-fitting, the stability score attains its maximum value.

To overcome this drawback, a new stability measure employing an adjustment $\text{Adj}(V_i, V_j)$ different from $\frac{A(V_i, V_j) + A(V_j, V_i)}{2}$ that fulfills \[\max\left[ \labs V_i \cap V_j \rabs + \text{Adj}(V_i, V_j)\right] \leq \max\left[\labs \widetilde{V}_i \cap \widetilde{V}_j \rabs \right] \text{ with } \labs \widetilde{V}_i \rabs = \labs V_i \rabs \text{ and } \labs \widetilde{V}_j \rabs = \labs V_j \rabs\] is defined in this paper.
This means that the adjusted score for $V_i$ and $V_j$ cannot exceed the value of $| \widetilde{V}_i \cap \widetilde{V}_j |$ that would be obtained if two sets $\widetilde{V}_i$ and $\widetilde{V}_j$ with $| \widetilde{V}_i | = | V_i |$ and $ | \widetilde{V}_j | = | V_j |$ were chosen such that their overlap is maximal.
This happens when $\widetilde{V}_i \subseteq \widetilde{V}_j$ or $\widetilde{V}_j \subseteq \widetilde{V}_i$.
The resulting measure is
\[\text{SMA} = \frac{2}{m (m-1)} \sum\limits_{i=1}^{m-1} \sum\limits_{j = i+1}^m \frac{\labs V_i \cap V_j \rabs + \text{Adj}(V_i, V_j) - E\left[ \labs V_i \cap V_j \rabs + \text{Adj}(V_i, V_j)\right]}{\text{UB}\left[ \labs V_i \cap V_j \rabs \right] - E\left[ \labs V_i \cap V_j \rabs + \text{Adj}(V_i, V_j)\right]}\]
with $\text{UB}\left[ \labs V_i \cap V_j \rabs \right]$ denoting an upper bound for $\labs V_i \cap V_j \rabs$.
The expected values $E\left[ \labs V_i \cap V_j \rabs + \text{Adj}(V_i, V_j)\right]$ cannot be calculated with a universal formula as they depend on the data specific similarity structure.
However, they can be estimated by repeating the following Monte-Carlo-procedure $N$~times:
1.~Randomly draw sets $\widetilde{V}_i \subseteq \{X_1, \ldots, X_p\}$ and $\widetilde{V}_j \subseteq \{X_1, \ldots, X_p\}$, with $| \widetilde{V}_i | = | V_i |$,  $\| \widetilde{V}_j | = | V_j |$, and equal selection probabilities for all features.
2.~Calculate the score $| \widetilde{V}_i \cap \widetilde{V}_j | + \text{Adj}(\widetilde{V}_i, \widetilde{V}_j)$.
An estimate for the expected value $E \left[ \labs V_i \cap V_j \rabs + \text{Adj}(V_i, V_j) \right]$ is the average of the $N$ scores.

Concerning the upper bounds $\text{UB} \left[ \labs V_i \cap V_j \rabs \right]$, $\min\{\labs V_i \rabs, \labs V_j \rabs \}$ is the tightest upper bound for $\labs V_i \cap V_j \rabs$.
However, this upper bound is not a good choice for $\text{UB} \left[ \labs V_i \cap V_j \rabs \right]$ because the stability measure could attain its maximum value for sets $V_i \subsetneqq V_j$ or $V_j \subsetneqq V_i$.
To avoid it, $\text{UB} \left[ \labs V_i \cap V_j \rabs \right]$ must depend on both $\labs V_i \rabs$ and $\labs V_j \rabs$.
Possible choices are for example $\frac{\labs V_i \rabs + \labs V_j \rabs}{2}$ or $\sqrt{\labs V_i\rabs \cdot \labs V_j \rabs}$.
These choices are upper bounds for $\labs V_i \cap V_j \rabs$ and they are met if and only if $V_i = V_j$.
For $\labs V_i\rabs \neq \labs V_j \rabs$, the bounds differ and $\min\left\{ \labs V_i \rabs, \labs V_j \rabs \right\} \leq \sqrt{ \labs V_i \rabs \cdot \labs V_j \rabs} \leq \frac{ \labs V_i \rabs + \labs V_j \rabs}{2}$ holds which makes $\sqrt{ \labs V_i\rabs \cdot \labs V_j \rabs}$ more suitable.
Therefore, $\text{UB} \left[ \labs V_i \cap V_j \rabs \right] = \sqrt{ \labs V_i \rabs \cdot \labs V_j \rabs}$ is used in this paper.
If there are no similar features in the data set, SMA is identical to SMU, independent of the choice of adjustment.
Four different adjustments are considered.
We first define them and then give explanations for their construction.

\begin{align*}
\text{Adj}_{\text{MBM}}(V_i, V_j) &= \text{size of maximum bipartite matching} \, (V_i \setminus V_j, V_j \setminus V_i)\\
\text{Adj}_{\text{Greedy}}(V_i, V_j) &= \text{greedy choice of most similar pairs of features} \\
&\quad \text{determined by Algorithm~\ref{algo.greedy} introduced on page \pageref{algo.greedy}}\\
\text{Adj}_{\text{Count}}(V_i, V_j) &= \min\{ A(V_i, V_j), A(V_j, V_i)\}
\text{ with A as defined for SMY}\\
\text{Adj}_{\text{Mean}}(V_i, V_j) &= \min\{ M(V_i, V_j), M(V_j, V_i)\}
\text{ with}\\
&\quad M(V_i, V_j) = \sum\limits_{x \in  V_i \setminus V_j : \labs G_x^{ij} \rabs > 0} \frac{1}{\labs G_x^{ij} \rabs} \sum \limits_{y \in G_x^{ij}} \text{similarity}(x,y) \text{ and}\\
&\quad G_x^{ij} = \left\{ y \in V_j \setminus V_i : \text{similarity}(x,y) \geq \theta \right\} 
\end{align*}
The resulting four variants of SMA are named SMA-MBM, SMA-Greedy, SMA-Count and SMA-Mean.
For the adjustment of SMA-MBM, a graph is constructed.
In this graph, each feature of $(V_i \setminus V_j) \cup (V_j \setminus V_i)$ is represented by a vertex.
Vertices $x \in V_i \setminus V_j$ and $y \in V_j \setminus V_i$ are connected by an edge, if and only if similarity$(x,y) \geq \theta$.
An edge in the graph means that the corresponding features of the two connected vertices should be seen as exchangeable for stability assessment.
A matching of a graph is defined as a subset of its edges such that none of the edges share a vertex \cite[p.~63]{rahman2017basic}.
A maximum matching is a matching that contains as many edges as possible.
The size of the maximum matching is the number of edges that are included in the maximum matching.
The size of the maximum matching can be interpreted as the maximum number of features in $V_i \setminus V_j$ and $V_j \setminus V_i$ that should be seen as exchangeable for stability assessment with the restriction that each feature in $V_i \setminus V_j$ may only be seen as exchangeable with at most one feature in $V_j \setminus V_i$ and vice versa.
There are no edges between vertices that both correspond to features of $V_i \setminus V_j$ or $V_j \setminus V_i$, so the graph is bipartite \cite[p.~17]{rahman2017basic}.
For the calculation of a maximum matching for a bipartite graph, there exist specific algorithms \cite{hopcroft1973n}.

The calculation of the maximum bipartite matching has the complexity $\mathcal{O}((\text{number of vertices} + \text{number of edges}) \cdot \sqrt{\text{number of vertices}})$ \cite{hopcroft1973n} and hence can be very time consuming.
Therefore, a new greedy algorithm for choosing the most similar pairs of features is introduced in Algorithm~\ref{algo.greedy}.
It is used to calculate the adjustment in SMA-Greedy.
The return value of the algorithm is always smaller than or equal to the size of the maximum bipartite matching of the corresponding graph.
The computational complexity of the algorithm is dominated by the sorting of the edges and hence is $\mathcal{O}\left(\text{number of edges} \cdot \log(\text{number of edges})\right)$.
\begin{algorithm}[tb]
\DontPrintSemicolon
\caption{Greedy choice of the most similar pairs of features.} \label{algo.greedy}
size = 0\;
$L_A = [X, Y, S]$ = list of tuples $x \in V_i \setminus V_j$, $y \in V_j \setminus V_i$, similarity$(x,y)$ with similarity$(x,y) \geq \theta$, sorted decreasingly by similarity values\;
$L_B$ = empty list\;
\While{length of $L_A > 0$}{
$[x, y, s]$ = first tuple of $L_A$\;
add $[x, y, s]$ to $L_B$\;
remove all tuples in $L_A$ that contain $x$ or $y$\;
}
\Return{length of $L_B$}
\end{algorithm} 

For SMA-Count, $A(V_i, V_j)$ is the number of features in $V_i$, that are not in $V_j$ but that have a similar feature in $V_j \setminus V_i$.
The minimum of $A(V_i, V_j)$ and $A(V_j, V_i)$ is used in order to guarantee that the adjusted score for $V_i$ and $V_j$ cannot exceed the value of $ | \widetilde{V}_i \cap \widetilde{V}_j |$ that would be obtained if two sets $\widetilde{V}_i$ and $\widetilde{V}_j$ with $| \widetilde{V}_i | = | V_i |$ and $| \widetilde{V}_j | = | V_j |$ were chosen such that their overlap is maximal.
$\min\{A(V_i, V_j), A(V_j, V_i)\}$ is always larger than or equal to the size of the maximum bipartite matching.

The adjustment of SMA-Mean is very similar to the one of SMA-Count.
While $A(V_i, V_j)$ counts the number of features in $V_i \setminus V_j$, that have a similar feature in $V_j \setminus V_i$, $M(V_i, V_j)$ sums up the mean similarity values of the features in $V_i \setminus V_j$ to their similar features in $V_j \setminus V_i$.
If there are no similarity values of features in $V_i \setminus V_j$ and $V_j \setminus V_i$ in the interval $[\theta, 1)$, the adjustments of SMA-Count and SMA-Mean are identical.
Otherwise, the adjustment of SMA-Mean is smaller than the adjustment of SMA-Count.

\section{Experiments and Results}\label{sec.experiments.results}
The adjusted stability measures SMZ, SMY, SMA-Count, SMA-Mean, SMA-Greedy and SMA-MBM are compared to each other and to the unadjusted measure SMU.
All calculations have been performed with the software \textit{R}~\cite{R} using the package \textit{stabm}~\cite{stabm} for calculating the stability measures and \textit{batchtools}~\cite{batchtools} for conducting the experiments on a high performance compute cluster.

\subsection{Experimental Results on Artificial Feature Sets}

\begin{figure}[tb]
\centerline{\includegraphics[scale = 0.45]{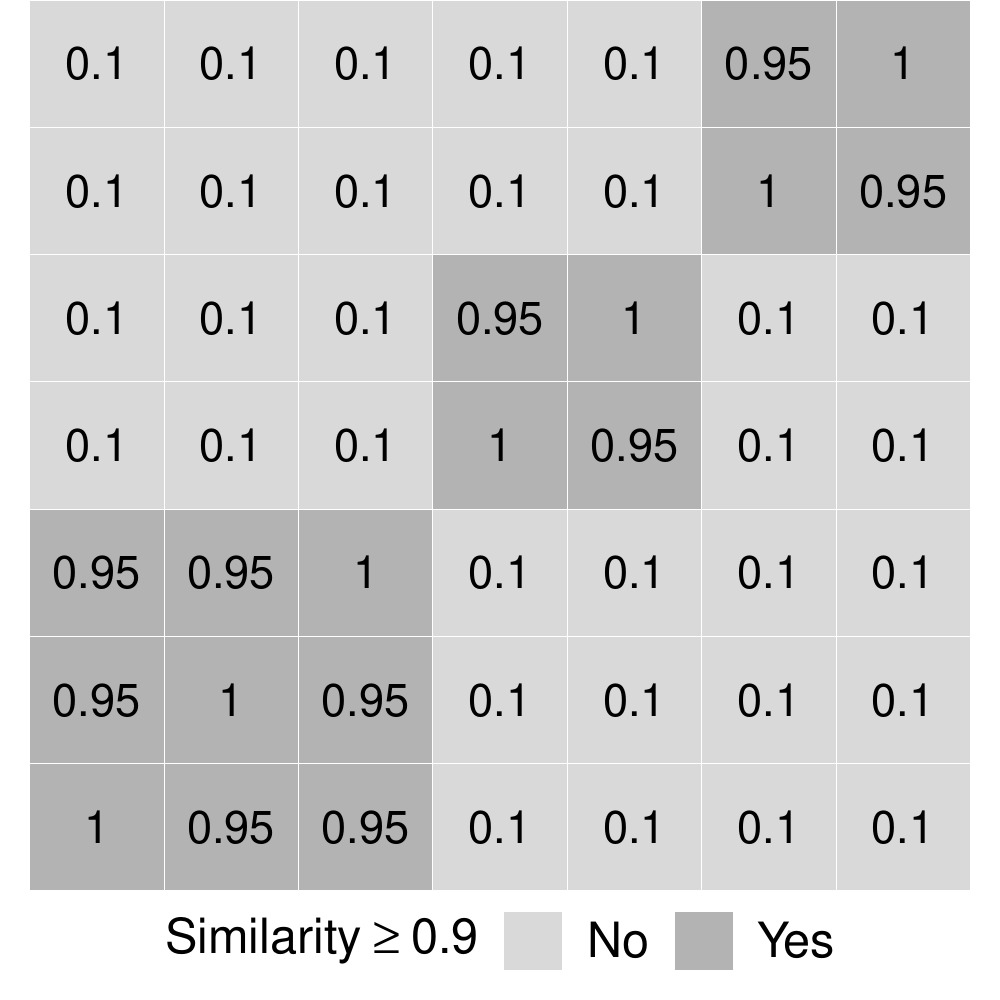}}
\caption{Similarity matrix for the 7 features. Similarity values must be in $[0, 1]$.}
\label{fig.simmat}
\end{figure}

First, a comparison in a situation with only 7~features is conducted. 
The advantage of this comparison is that all possible combinations
of 2~subsets of features can be analyzed, as there are only $2^7 \cdot 2^7 = 16\,384$ possible combinations. 
For the adjusted and corrected measures SMY, SMA-Count, SMA-Mean, SMA-Greedy and SMA-MBM, the expected values of the pairwise scores are calculated exactly by considering all possible pairs of sets of the same cardinalities.
The values of all stability measures presented in Subsection~\ref{sec.measures} are calculated for all 16\,384 possible combinations of 2~feature sets being selected from a total number of 7~features.
Figure~\ref{fig.simmat} displays the similarities between the 7~features  used for this analysis.
The threshold $\theta$ is set to $\theta = 0.9$, so there are 3~groups of similar features.
Note that the similarity matrix is sufficient for calculating the stability measure for all pairs of possible combinations of 2 feature sets.

\begin{figure}[tb]
\centerline{\includegraphics[width = 0.9\textwidth]{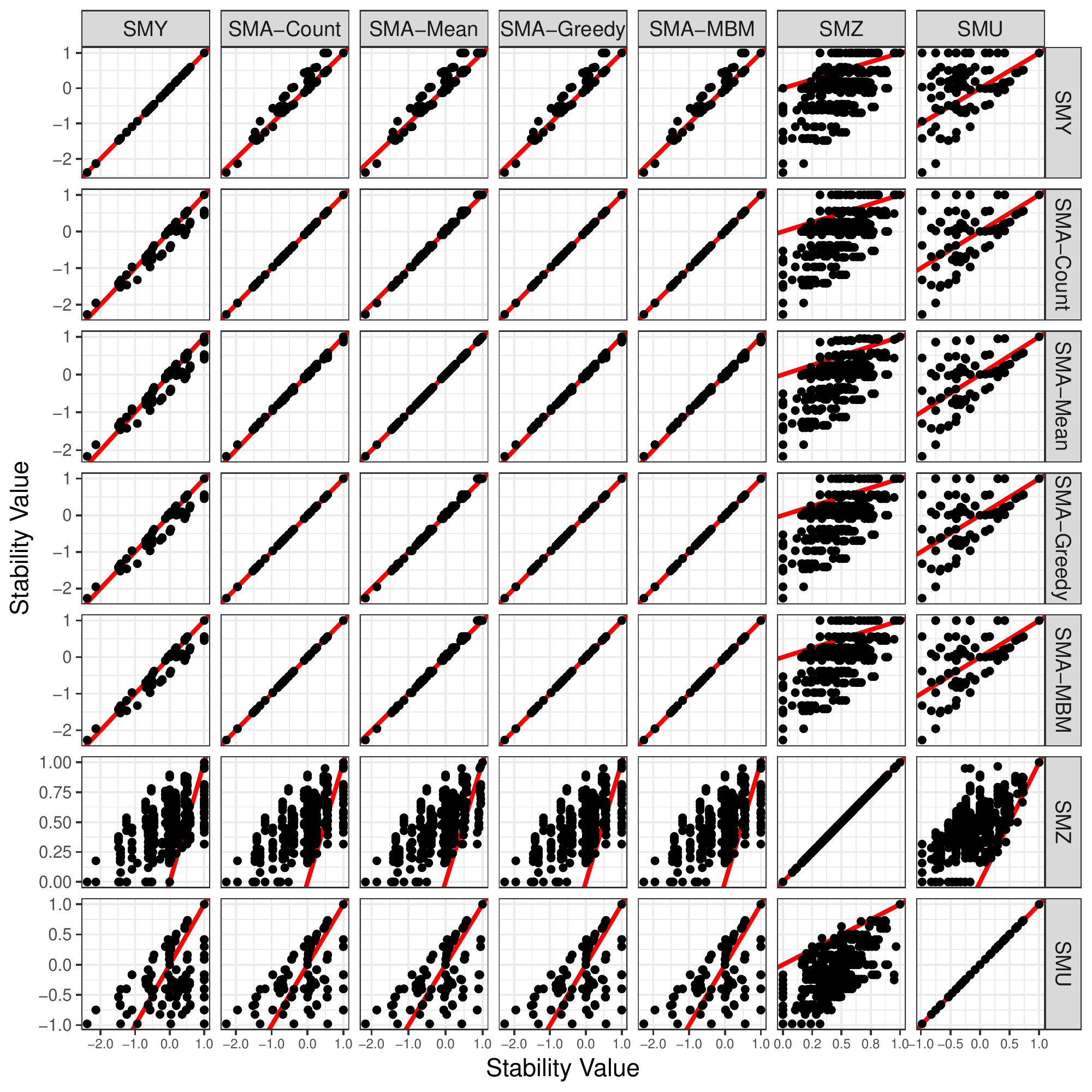}}
\caption{Scatter plots of the stability values for the 16\,384 combinations and all seven stability measures.
The line in each plot indicates the identity.}
\label{fig.scatter}
\end{figure}

To compare all stability measures, in Figure~\ref{fig.scatter}, scatter plots of all pairs of stability measures are shown.
All adjusted measures differ strongly from the unadjusted stability measure SMU with respect to their stability assessment behavior.
The adjusted stability measure SMZ, which is the only considered measure that is not corrected for chance, also differs strongly from all other stability measures.
This demonstrates, that the missing correction has a large impact on the stability assessment behavior.
SMA-Count, SMA-Mean, SMA-Greedy and SMA-MBM have almost identical values for all combinations.
The values assigned by SMY and by the SMA variants are also quite similar.
However, for combinations that obtain comparably large stability values by all of these measures, SMY often attains larger values than the SMA measures.
These are combinations for which several features from the one set are mapped to the same feature of the other set, see the discussion in Subsection~\ref{sec.measures}.
This undesired behavior of SMY occurs for large stability values.
This is problematic because large stability values are what an optimizer is searching for when fitting models in a multi-criteria fashion taking into account the feature selection stability~\cite{bommert2017multicriteria}.

\subsection{Experimental Results on Real Feature Sets}
Now, the stability measures are compared based on feature sets that are selected for four real data sets with correlated features (OpenML~\cite{vanschoren2013openml} IDs 851, 41\,163, 1\,484 and 1\,458) with  feature selection methods.
The details of the feature selections are omitted here due to space constraints.
Also, the focus is on the evaluation of the stability based on realistic feature sets resulting from real applications.
To assess the similarity between features, the absolute Pearson correlation is employed for all adjusted stability measures. 
The threshold $\theta$ is set to $\theta = 0.9$ because in many fields, an absolute correlation of 0.9 or more is interpreted as a \enquote{strong} or even \enquote{very strong} association.
For the adjusted and corrected measures SMY, SMA-Count, SMA-Mean, SMA-Greedy and SMA-MBM, the expected values of the pairwise scores are estimated based on $N = 10\,000$ replications.
This value for $N$ is suggested in~\cite{zhang2009evaluating} and has shown to provide a good compromise between convergence and run time in preliminary studies.

\begin{figure}[htb]
\centerline{\includegraphics[scale = 0.55]{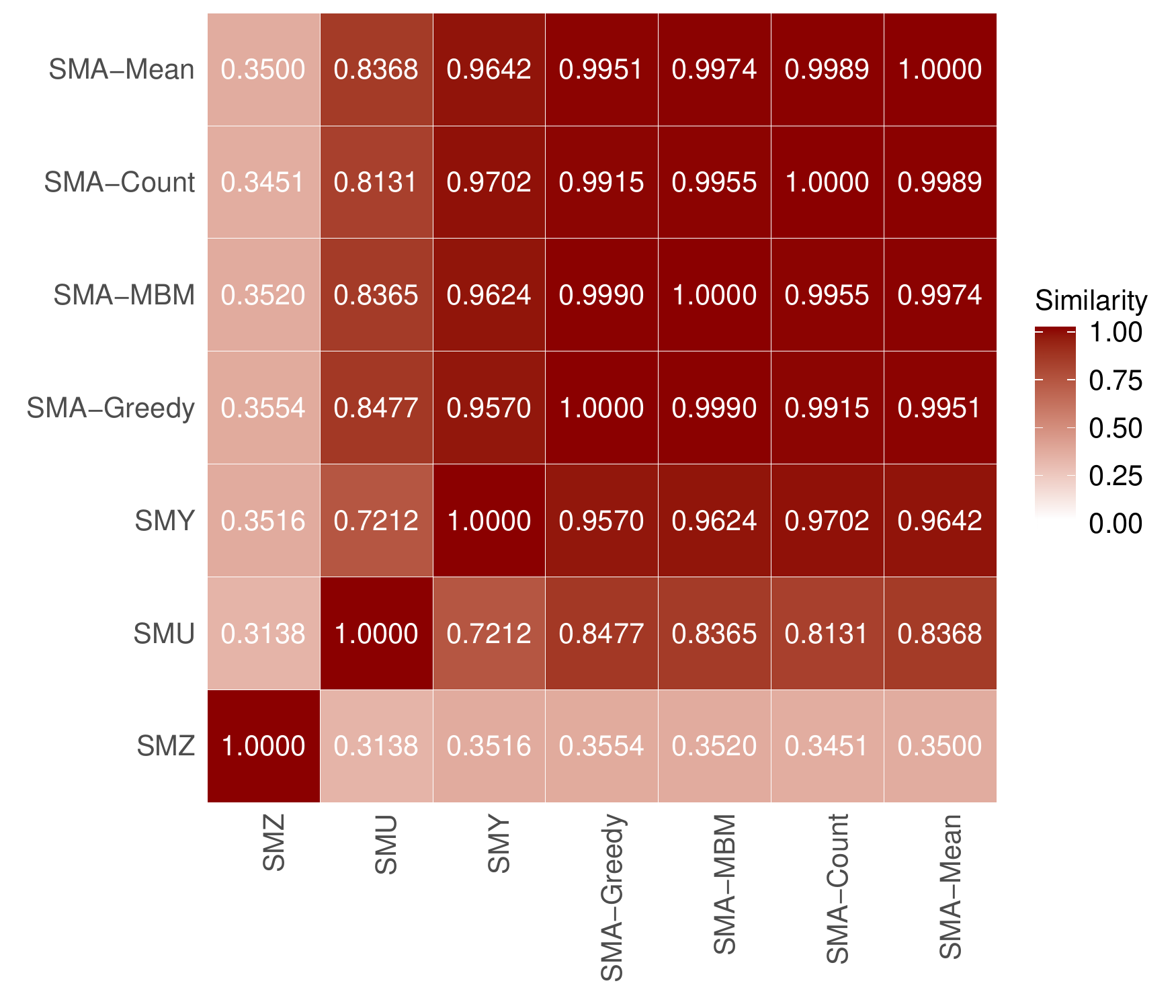}}
\caption{Mean Pearson correlations between all pairs of the seven stability measures.
The correlations between the stability measures are averaged across data sets.}
\label{fig.cors}
\end{figure}

To analyze the similarities between the stability measures, Pearson correlations between all pairs of stability measures are calculated and averaged across data sets by calculating the arithmetic mean.
Figure~\ref{fig.cors}
displays the results.
The adjusted and uncorrected stability measure SMZ differs most strongly from all other stability measures.
The adjusted and corrected measures SMY, SMA-Count, SMA-Mean, SMA-Greedy and SMA-MBM assess the stability almost identically.
The corrected and unadjusted measure SMU is more similar to this group  than to SMZ.
Here, SMU is much more similar to the corrected and adjusted stability measures SMY, SMA-Count, SMA-Mean, SMA-Greedy and SMA-MBM than in the previous subsection.
The reason is that the real data sets considered here contain fewer similar features in comparison to the total number of features than in the artificial example in the previous subsection.

\begin{figure}[tb]
\centerline{\includegraphics[width = \textwidth]{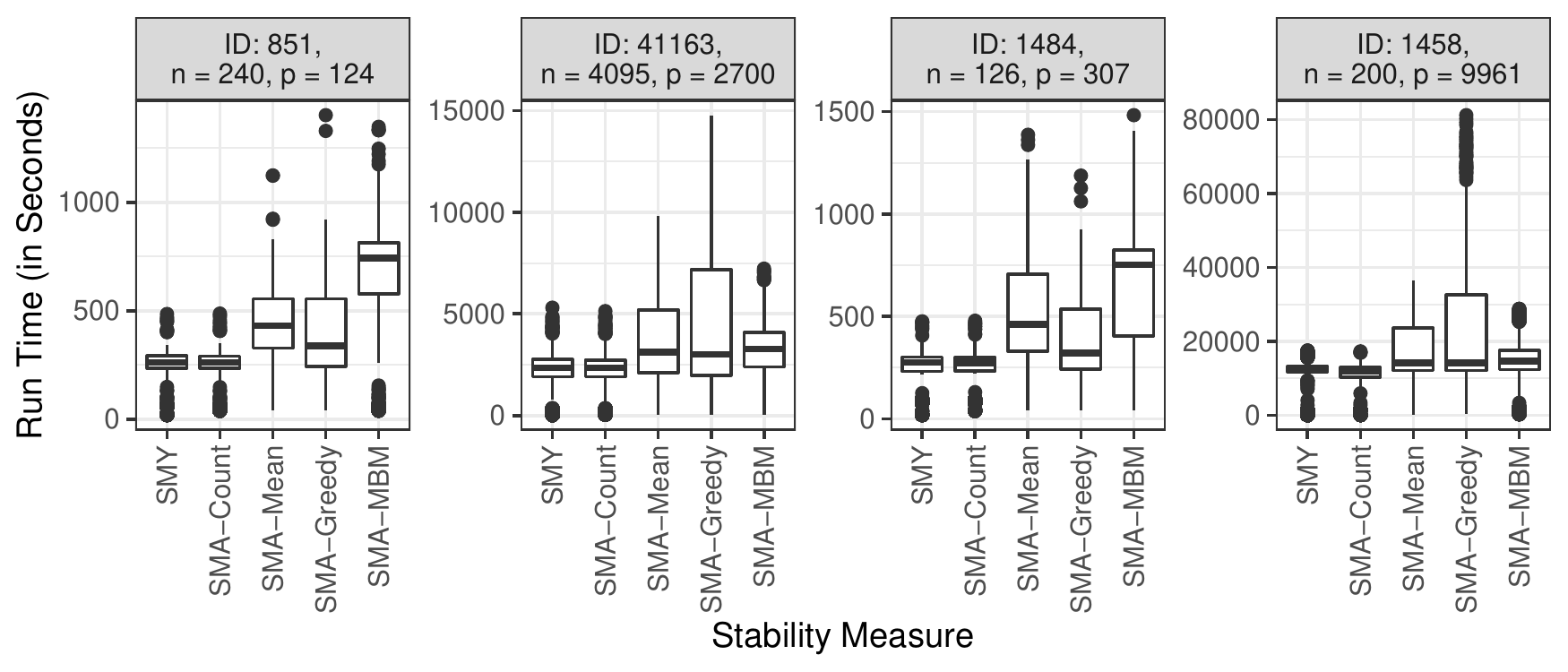}}
\caption{Run times of the adjusted and corrected stability measures for four real data sets.
$n$:~number of observations, $p$:~number of features.}
\label{fig.measures.runtimes.1}
\end{figure}

Now, the run times of the stability measures for realistic feature sets are compared.
For SMU and SMZ, the run time is not an issue.
For all of the considered data sets, they can be computed in less than one second.
Figure~\ref{fig.measures.runtimes.1}
displays the run times for calculating the values of the adjusted and corrected stability measures.
The run times of these measures are much longer than the run times of SMU and SMZ.
The reason is that the expected values of the scores have to be estimated, which involves frequently repeated evaluation of the adjustments.
For all data sets, SMY and SMA-Count require the least time for calculation among the adjusted and corrected measures.
For most data sets, the calculation of SMA-Mean, SMA-Greedy and SMA-MBM takes much longer.
For large data sets, the latter computation times are not acceptable.

\section{Conclusions}\label{sec.conclusions}
For data sets with similar features, for example data sets with highly correlated features, the evaluation of feature selection stability is difficult.
The commonly used stability measures consider features that are almost identical but have different identifiers as different features.
This, however, is not desired because almost the same information is captured by the respective sets of features.

We have introduced and investigated new stability measures that take into account similarities between features (\enquote{adjusted} stability measures).
We have compared them to existing stability measures based on both artificial and real sets of selected features.
For the existing stability measures, drawbacks were explained and demonstrated.

For the newly proposed adjusted stability measure SMA, four variants were considered: SMA-Count, SMA-Mean, SMA-Greedy and SMA-MBM.
They differ in the way they take into account similar features when evaluating the stability.
Even though the adjustments for similar features are conceptually different for the four variants, the results are very similar both on artificial and on real sets of selected features.
With respect to run time, the variant SMA-Count outperformed the others.
Therefore, we conclude that SMA-Count should be used when evaluating the feature selection stability for data sets with similar features.

A promising future strategy is to employ SMA-Count when searching for models with high predictive accuracy, a small number of chosen features and a stable feature selection for data sets with similar features.
To reach this goal, one can perform multi-criteria hyperparameter tuning with respect to the three criteria and assess the stability with SMA-Count.

\section*{Acknowledgements}
This work was supported by German Research Foundation (DFG), Project RA\,870/7-1 and Collaborative Research Center SFB~876, A3. We acknowledge the computing time provided on the Linux HPC cluster at TU Dortmund University (LiDO3), partially funded in the course of the Large-Scale Equipment Initiative by the German Research Foundation (DFG) as Project 271512359.

\bibliographystyle{splncs04}
\bibliography{literature}

\begin{thebibliography}{10}
\providecommand{\url}[1]{\texttt{#1}}
\providecommand{\urlprefix}{URL }
\providecommand{\doi}[1]{https://doi.org/#1}

\bibitem{awada2012review}
Awada, W., Khoshgoftaar, T.M., Dittman, D., Wald, R., Napolitano, A.: A review
  of the stability of feature selection techniques for bioinformatics data. In:
  2012 IEEE International Conference on Information Reuse and Integration. pp.
  356--363 (2012)

\bibitem{stabm}
Bommert, A.: stabm: Stability Measures for Feature Selection (2019),
  \url{https://CRAN.R-project.org/package=stabm}, {R} package version 1.1.0

\bibitem{bommert2017multicriteria}
Bommert, A., Rahnenf{\"u}hrer, J., Lang, M.: A multicriteria approach to find
  predictive and sparse models with stable feature selection for
  high-dimensional data. Computational and Mathematical Methods in Medicine
  \textbf{2017}, 7907163 (2017)

\bibitem{he2010stable}
He, Z., Yu, W.: Stable feature selection for biomarker discovery. Computational
  Biology and Chemistry  \textbf{34}(4),  215--225 (2010)

\bibitem{hopcroft1973n}
Hopcroft, J.E., Karp, R.M.: An {$n^{5/2}$} algorithm for maximum matchings in
  bipartite graphs. SIAM Journal on Computing  \textbf{2}(4),  225--231 (1973)

\bibitem{jaccard1901etude}
Jaccard, P.: {\'E}tude comparative de la distribution florale dans une portion
  des {Alpes} et du {Jura}. Bulletin de la Soci{\'e}t{\'e} Vaudoise des
  Sciences Naturelles  \textbf{37},  547--579 (1901)

\bibitem{kalousis2007stability}
Kalousis, A., Prados, J., Hilario, M.: Stability of feature selection
  algorithms: A study on high-dimensional spaces. Knowledge and Information
  Systems  \textbf{12}(1),  95--116 (2007)

\bibitem{batchtools}
Lang, M., Bischl, B., Surmann, D.: batchtools: Tools for {R} to work on batch
  systems. Journal of Open Source Software  \textbf{2}(10) (2017)

\bibitem{lausser2013measuring}
Lausser, L., M{\"u}ssel, C., Maucher, M., Kestler, H.A.: Measuring and
  visualizing the stability of biomarker selection techniques. Computational
  Statistics  \textbf{28}(1),  51--65 (2013)

\bibitem{nogueira2018quantifying}
Nogueira, S.: Quantifying the Stability of Feature Selection. Ph.D. thesis,
  University of Manchester, United Kingdom (2018)

\bibitem{R}
{R Core Team}: {R}: A Language and Environment for Statistical Computing. R
  Foundation for Statistical Computing, Vienna, Austria (2018),
  \url{https://www.R-project.org/}

\bibitem{rahman2017basic}
Rahman, M.S.: Basic Graph Theory. Springer, New York, USA (2017)

\bibitem{statnikov2013algorithms}
Statnikov, A., Lytkin, N.I., Lemeire, J., Aliferis, C.F.: Algorithms for
  discovery of multiple markov boundaries. Journal of Machine Learning Research
   \textbf{14},  499--566 (2013)

\bibitem{vanschoren2013openml}
Vanschoren, J., Van~Rijn, J.N., Bischl, B., Torgo, L.: {OpenML}: Networked
  science in machine learning. ACM SIGKDD Explorations Newsletter
  \textbf{15}(2),  49--60 (2013)

\bibitem{yu2012stable}
Yu, L., Han, Y., Berens, M.E.: Stable gene selection from microarray data via
  sample weighting. IEEE/ACM Transactions on Computational Biology and
  Bioinformatics  \textbf{9}(1),  262--272 (2012)

\bibitem{zhang2009evaluating}
Zhang, M., Zhang, L., Zou, J., Yao, C., Xiao, H., Liu, Q., Wang, J., Wang, D.,
  Wang, C., Guo, Z.: Evaluating reproducibility of differential expression
  discoveries in microarray studies by considering correlated molecular
  changes. Bioinformatics  \textbf{25}(13),  1662--1668 (2009)

\bibitem{zucknick2008comparing}
Zucknick, M., Richardson, S., Stronach, E.A.: Comparing the characteristics of
  gene expression profiles derived by univariate and multivariate
  classification methods. Statistical Applications in Genetics and Molecular
  Biology  \textbf{7}(1), 7 (2008)

\end{thebibliography}

\end{document}